%% file: main.tex
\documentclass[letterpaper]{article} % DO NOT CHANGE THIS
\usepackage{aaai24}  % DO NOT CHANGE THIS
\usepackage{times}  % DO NOT CHANGE THIS
\usepackage{helvet}  % DO NOT CHANGE THIS
\usepackage{courier}  % DO NOT CHANGE THIS
\usepackage[hyphens]{url}  % DO NOT CHANGE THIS
\usepackage{graphicx} % DO NOT CHANGE THIS
\usepackage{natbib}  % DO NOT CHANGE THIS AND DO NOT ADD ANY OPTIONS TO IT
\usepackage{caption} % DO NOT CHANGE THIS AND DO NOT ADD ANY OPTIONS TO IT
\usepackage{algorithm}
\usepackage{algorithmic}
\usepackage{pgfplots}
\usepackage{color}

\def\ie{\textit{i.e.}}
\def\eg{\textit{e.g.}}
\usepackage{newfloat}
\usepackage{listings}
\usepackage{caption}
\usepackage{subcaption}
\usepackage{graphicx}
\usepackage{multirow}
\usepackage{makecell}
\usepackage{booktabs}
\usepackage{pgfplots}
\pgfplotsset{width=6cm,compat=1.9}
\usepackage{amsmath,amssymb}
\usepackage{bm}
\usepackage{arydshln}
\usepackage{booktabs}

\def\ie{\textit{i.e.}}
\def\eg{\textit{e.g.}}

\definecolor{nmgray}{RGB}{229,229,229}
\definecolor{underlinegray}{RGB}{197,197,197}

\definecolor{introblue}{RGB}{0,176,240}
\definecolor{introgreen}{RGB}{0,203,134}
\definecolor{introgreen2}{RGB}{139,243,206}

\usepackage{tcolorbox}

\usepackage{adjustbox}
\newtcolorbox{mybox}[2][]{
width=\columnwidth,
colback = nmgray!75!white, 
colframe = nmgray!75!white, 
boxsep=0pt,left=10pt,right=10pt,top=0pt,bottom=0pt,
fontupper=\linespread{0.9}\selectfont,
title=#2,#1}

\newcommand{\vpara}[1]{\vspace{0.05in}\noindent \textbf{#1 }}

\DeclareCaptionStyle{ruled}{labelfont=normalfont,labelsep=colon,strut=off} % DO NOT CHANGE THIS
\floatstyle{ruled}
\newfloat{listing}{tb}{lst}{}
\floatname{listing}{Listing}

\title{Reverse Multi-Choice Dialogue Commonsense Inference with Graph-of-Thought}

\author{
    Li Zheng\textsuperscript{\rm 1},
    Hao Fei\textsuperscript{\rm 2},
    Fei Li\textsuperscript{\rm 1}\thanks{Corresponding author.},
    Bobo Li\textsuperscript{\rm 1},
    Lizi Liao\textsuperscript{\rm 3},
    Donghong Ji\textsuperscript{\rm 1},
    Chong Teng\textsuperscript{\rm 1}
}
\affiliations{
    \textsuperscript{\rm 1}Key Laboratory of Aerospace Information Security and Trusted Computing, \\Ministry of Education, School of Cyber Science and Engineering, Wuhan University, Wuhan, China\\
    \textsuperscript{\rm 2}National University of Singapore \quad \textsuperscript{\rm 3}Singapore Management University
    \texttt{zhengli@whu.edu.cn, haofei37@nus.edu.sg, foxlf823@gmail.com, boboli@whu.edu.cn,} \\\texttt{lzliao@smu.edu.sg, dhji@whu.edu.cn, tengchong@whu.edu.cn}
}

\usepackage{bibentry}
\begin{document}
% \begin{CJK}{UTF8}{gbsn}
\maketitle

\begin{abstract}

With the proliferation of dialogic data across the Internet, the \textit{Dialogue Commonsense Multi-choice Question Answering} (DC-MCQ) task has emerged as a response to the challenge of comprehending user queries and intentions.
Although prevailing methodologies exhibit effectiveness in addressing single-choice questions, they encounter difficulties in handling multi-choice queries due to the heightened intricacy and informational density. 
In this paper, inspired by the human cognitive process of progressively excluding options, we propose a three-step \emph{Reverse Exclusion Graph-of-Thought} (\textbf{ReX-GoT}) framework, including Option Exclusion, Error Analysis, and Combine Information.
Specifically, our ReX-GoT mimics human reasoning by gradually excluding irrelevant options and learning the reasons for option errors to choose the optimal path of the GoT and ultimately infer the correct answer.
By progressively integrating intricate clues, our method effectively reduces the difficulty of multi-choice reasoning and provides a novel solution for DC-MCQ.
Extensive experiments on the CICERO and CICERO$_{v2}$ datasets validate the significant improvement of our approach on DC-MCQ task.
On zero-shot setting, our model outperform the best baseline by 17.67\% in terms of F1 score for the multi-choice task.
Most strikingly, our GPT3.5-based ReX-GoT framework achieves a remarkable 39.44\% increase in F1 score.
Our code is available at: \url{https://github.com/ZhengL00/ReX-GoT}.

\end{abstract}

\input{Section/1-introduction}

\input{Section/2-relatedwork}

\input{Section/3-methodology}

\input{Section/4-experiments}
\input{Section/5-conclusion}

%\bibliography{aaai24}

%\appendix

\section*{Acknowledgements}
This work is supported by the National Key Research and Development Program of China (No. 2022YFB3103602), the National Natural Science Foundation of China (No. 62176187), the open project of Sichuan Provincial Key Laboratory of Philosophy, the Social Science for Language Intelligence in Special Education (No. YYZN-2023-1) and CCF-Baidu Open Fund.

\bibliography{aaai24}

% \end{CJK}
\end{document}

%% file: Section/1-introduction.tex
\section{Introduction}
Commonsense knowledge is crucial for human cognition and natural human-computer interactions, which encompasses our intuitive understanding of the world and ability to reason. 
With the growth of social networks, commonsense inference \cite{DBLP:conf/aaai/ArabshahiLGMAM21,DBLP:conf/acl/0010LLWWBCH22,DBLP:conf/acl/KuoC23} in dialogue has garnered noteworthy attention as a burgeoning research domain in natural language processing (NLP).
However, accurately understanding and interpreting speaker questions and intentions in dialogue poses an essential challenge. 
To this end, the \textit{Dialogue Commonsense Multi-choice Question Answering} task~\cite{ghosal2022cicero} was proposed, defined as to select logical answers from preset options based on dialogue's history and context.

 \begin{figure}[!t]
    \centering
    \includegraphics[scale=0.72]{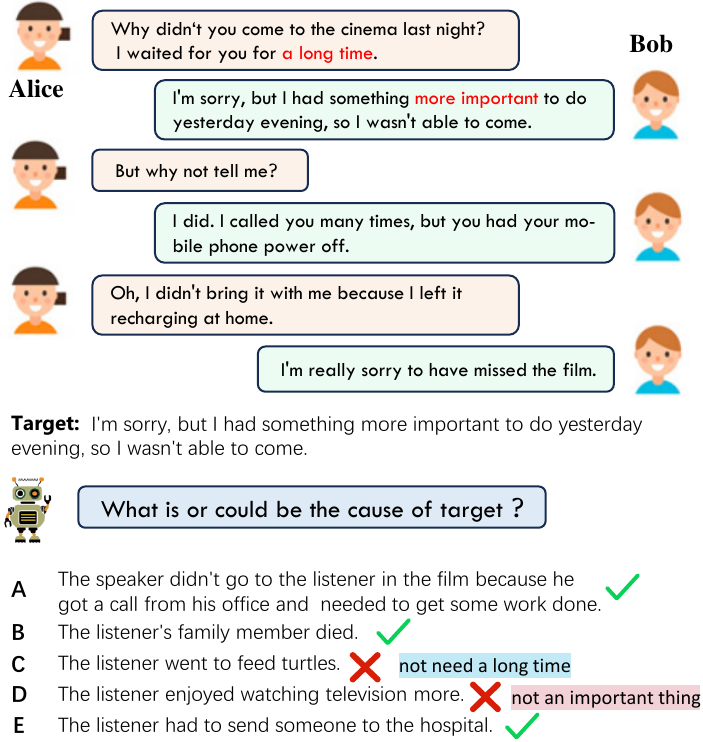}
    \caption{An example from the CICERO$_{v_2}$ dataset about dialogue commonsense reasoning.
    }
    \label{fig:demo}
    \vspace{-0.4cm}
\end{figure}

DC-MCQ task involve both single-choice and multi-choice questions. 
While existing works~\cite{wang2018co, zhang2020dcmn+,ju2021enhancing} achieved promising results in single-choice task, the performance in multi-choice task remains unsatisfactory.
Due to the intricate nature of multi-choice task, the challenges of \textbf{Option Saturation} and \textbf{Clue Labyrinth} burden current models. 
The option saturation challenge refers to the uncertainty of the number of options, which increases the difficulty of inference for the model. 
On a parallel note, the clue labyrinth challenge involves analyzing the combination of different complex clues, which includes intricate hidden information woven throughout the question stem and answer options, and different clues of predicted information, just like the complexity of the labyrinth.
It demands enhanced information integration comprehension by the model.
Hence, multi-choice questions are significantly more challenging than single-choice ones. 
As indicated by~\citet{ghosal2022two}, the community recognizes that attaining high accuracy with such questions is a potentially insurmountable task.

Existing methods for multi-choice questions, as highlighted by \citet{ghosal2022cicero} and \citet{shen2022multiview}, predominantly rely on forward reasoning. 
This typically assesses each option in isolation, which falters in accurately identifying the right answers due to intricate interrelations and uncertainties among choices.
Motivated by human cognitive patterns of option exclusion, we employ a similar tactic to progressively narrow down potential answers. 
As exemplified in Figure \ref{fig:demo}, depending on the context, we exclude certain options such as D and C, obtaining clues that Bob had something more important to do and the correct option must being important as well as taking a long time. 
Continuing reasoning based on the context and the clues we have, we determine that options A, B, and E are correct.
This exclusion-centric approach enhances reasoning, uncovers obscured insights from incorrect options, and greatly eases the prediction challenge in multi-answer scenarios.

On the other hand, the context of each option in the multi-choice task is broad enough to go beyond the scope of the given dialogue. 
Models based on direct answer selection struggle to fully comprehend the multi-dimensional and complex relationships between the question and the options, which can lead to model reasoning overload affecting accuracy.
With the widespread use of Large Language Models (LLMs) in NLP tasks, researchers~\cite{DBLP:conf/nips/Wei0SBIXCLZ22,fei2023reasoning,DBLP:conf/acl/JinL23,DBLP:conf/iclr/0001Z0S23} have identified the capacity of Chain-of-Thought (CoT) to help LLMs with complex reasoning tasks by generating intermediate steps.
However, the existing CoT reasoning of LLMs is limited to performing linear reasoning, and is unable to utilize potential multi-clue reasoning in a multi-dimensional manner to solve the clue labyrinth chanllenge.
Moreover, the existing CoT methods only superficially exploit the contextual information and overlook the utilization of the exclusion method to harness the hidden information within the options.

In this paper, based on the above observations, we design a three-step \emph{Reverse Exclusion Graph-of-Thought} (\textbf{ReX-GoT}) framework, including \textbf{Option Exclusion}, \textbf{Error Analysis}, and \textbf{Combine Information}.
ReX-GoT mimics human exclusion and selection methods by generating reverse exclusion graph-of-thought prompts. 
Concretely, leveraging LLMs as our basis, as shown in Figure~\ref{fig:method}, we initially prompt the model to discern irrational options and their underlying reasons. 
Subsequently, we utilize the insights gained in the first step for error analysis and option comparison to further guide the model to determine the rationality of each option and justifying its choice.
Finally, we combine the different reasons extracted in the first two steps as different paths and select the best path through a voting mechanism to arrive at the final multiple choice answer.
This distinctive amalgamation of backward exclusion and forward reasoning systematically excludes irrelevant alternatives and comprehends errors, thereby alleviating the complexity of predicting multiple correct responses.

To verify the effectiveness of our model, we conduct experiments on two widely-used datasets for DC-MCQ, namely CICERO \cite{ghosal2022cicero} and CICERO$_{v2}$ \cite{shen2022multiview}. 
On zero-shot setting, the experiment on the CICERO$_{v2}$ dataset show that the F1 score of our ReX-GoT is 17.67\% higher than the best baseline.
Most strikingly, our GPT3.5-based ReX-GoT with 175B
parameters boosts the baseline to a high-to 39.44\% increase of F1 score.

Our main contributions are summarized as follows:

\begin{itemize}
\item 

We first propose an reverse exclusion method that is consistent with human cognition, which effectively solves the challenge of option saturation by repeatedly excluding incongruous options and gradually revealing the hidden context of the correct option.
\item 

We design a brand-new GoT framework to productively address the clue labyrinth challenge. 
In this framework, different inference paths are set according to different analyses of the options, and the optimal path is finally selected to derive the correct answer.

\item Our extensive experimental results on CICERO and CIC-ERO$_{v2}$ datasets demonstrate that our scheme achieves state-of-the-art performance on the DC-MCQ task.

\end{itemize}

%% file: Section/2-relatedwork.tex
\section{Related Work}

\subsection{Commonsense Question Answering}

The domain of commonsense question answering  has garnered substantial attention within the realm of NLP. 
Existing models \cite{DBLP:conf/acl/ChenXYZHSZ23,DBLP:conf/aaai/DouP22,DBLP:conf/aaai/MaIFBNO21} have demonstrated remarkable capabilities in understanding and reasoning about common knowledge. 
Previous approaches such as prompt techniques~\cite{DBLP:conf/aaai/MaYLL23,DBLP:conf/acl/ZengWLF23,DBLP:conf/acl/ParanjapeMGHZ21} were proposed to improve the performance of language models in commonsense question answering task. 
Additionally, graph-based frameworks~\cite{DBLP:conf/acl/ZhaoHZSW23,DBLP:conf/nlpcc/ZhengLCTJ23,DBLP:conf/aaai/BosselutBC21}, including knowledge graphs and concept graphs, were also employed to enhance the representation and utilization of commonsense knowledge.

\subsection{Commonsense Inference in Dialogues}

Recently, there has been a growing interest in developing dialogue commonsense inference models \cite{DBLP:conf/aaai/ArabshahiLGMAM21,DBLP:conf/sigdial/GhosalHSMMP21,DBLP:journals/corr/abs-2302-07926}.
Several studies have been conducted on this topic:
\citet{DBLP:conf/acl/QinGUHCF20} investigated pre-trained language models for their temporal reasoning capabilities in dialogues. 
Furthermore, \citet{ghosal2022cicero} introduced a dialogue commonsense inference dataset CICERO 
that enables models to make educated guesses by considering the context when the answer is not obvious.
Shortly after, \citet{shen2022multiview} proposed the CICERO${v2}$ dataset, which provides more diverse options than CICERO.
Based on the two datasets, a recent study~\cite{ghosal2022two} transformed the task of selecting answers to a binary classification problem.
However, despite achieving certain results via direct classification, this approach overlooks the importance of step-by-step reasoning, which can significantly impact the analysis of results.

\subsection{LLM Reasoning with Chain-of-Thought}
The tremendous success of LLMs \cite{DBLP:conf/nips/BrownMRSKDNSSAA20,wu2023nextgpt} has propelled the development of various downstream applications, such as mathematical reasoning~\cite{DBLP:journals/corr/abs-2305-10601,DBLP:conf/acl/ImaniD023},
sentiment analysis~\cite{fei2023reasoning,DBLP:journals/corr/abs-2306-03969}, and chatbot~\cite{ouyang2022training,DBLP:journals/corr/abs-2307-08715}.
To exploit the reasoning ability in LLMs,
recent works~\cite{DBLP:conf/nips/Wei0SBIXCLZ22,DBLP:conf/acl/WangWLGYR23,DBLP:conf/acl/JinL23} started to explore the use of CoT in LLMs to enhance performance in complex tasks.
CoT prompting is an innovative gradient-free technique that guides LLMs to produce intermediate reasoning steps, ultimately leading to the derivation of the final answer.
Specifically, \citet{fei2023reasoning} introduce CoT into LLMs for implicit sentiment analysis.
\citet{DBLP:conf/acl/TrivediBKS23} interleave retrieval with steps in a CoT to improve question-answering performance.
More recently, \citet{DBLP:conf/acl/WangXLHLLL23} propose plan-and-solve prompting strategies to solve zero-shot CoT pitfalls.
Despite these recent advancements, LLMs with CoT have not been explored in dialogue commonsense inference.

%% file: Section/3-methodology.tex
\section{Methodology}
\label{sec:method}
\subsection{Task Definition}

The task of \textit{Dialogue Commonsense Multi-Choice Question Answering} (DC-MCQ) is defined as:
given a dialogue $D=\{u_1,... , u_n\}$, the target utterance $u_t$, for the target utterance commonsense question Q and the candidate options $F_t=\{f_{t_1},... , f_{t_m}\}$, a model selects all correct options $y$ in $F_t$.

\subsection{Preliminary}
\subsubsection{Standard Prompting}
\label{sec:prompt}

Standard prompting methods have been widely used in previous works~\cite{DBLP:conf/aaai/MaYLL23,DBLP:conf/acl/ParanjapeMGHZ21}. 
Through crafting specific prompts, LLMs can be fine-tuned to handle diverse tasks by simply changing the prompts
In this task, we construct the following prompt template as inputs for LLMs:

\vspace{-3pt}
\begin{mybox}\texttt
\texttt{Given the context T, which options are correct?}
\end{mybox}
\vspace{-3pt}
\noindent where $T = [D; u_t; Q; F_t]$, which includes dialogue, target utterance, question, and candidate options.

However, the prompting method has certain limitations. 
Firstly, it fails to account for option relationships, potentially resulting in erroneous predictions.
Secondly, the lack of explicit guidance for the LLMs to engage in a step-by-step reasoning process diminishes the interpretability of their answers.
As a result, comprehending the underlying logic behind a LLM's response becomes challenging.

\subsubsection{Vanilla CoT Prompting}
\label{sec:CoT}

To enhance the standard prompting method, chain-of-thought (CoT) prompting has been investigated, which advances in not only producing the answer, but eliciting LLMs to give the reasoning/rationale behind the answer.
For this task, we construct the following prompt template as inputs to LLMs:

\vspace{-3pt}
\begin{mybox}\texttt
\texttt{Given the context T, let's think step-by-step, which options are correct and why?}
\end{mybox}
\vspace{-3pt}

Nevertheless, the vanilla CoT merely directly prompts the model to generate intermediate inference processes and final results. 
While the existing CoT methods demonstrate some inference capabilities, they are limited to performing linear inference and fail to multidimensionally utilize multiple clues to reason about multiple options.
In addition, the vanilla CoT methods all direct the model to infer directly towards the answer in a forward manner, which easily overlook some of the correct options in cases with multiple valid answers, resulting in a performance decrease.
This approach does not align with the way humans typically approach multi-choice questions, which involves a combination of exclusion and forward reasoning.

\subsection{ReX-GoT Prompting}

Neither the aforementioned standard prompting nor the vanilla CoT approach can solve the option saturation and clue labyrinth challenges in DC-MCQ, so we propose a new approach called ReX-GoT, which stands for Reverse Exclusion with Graph-of-Thought. 
Our method leverages valuable information to guide the model to integrate clues for step-by-step reasoning in a reverse exclusion manner and in conjunction with a well designed GoT.
By doing so, our method effectively excludes incorrect options, narrows down the answer range, clarifies intricate clues, and improves the efficiency and accuracy of problem-solving.
Moreover, our method considers the logical relationships between the options and the context,  which contrasts with existing methods that solely rely on the contextual semantic information.

\begin{figure*}[!h]
    \centering
    \includegraphics[scale=0.55]{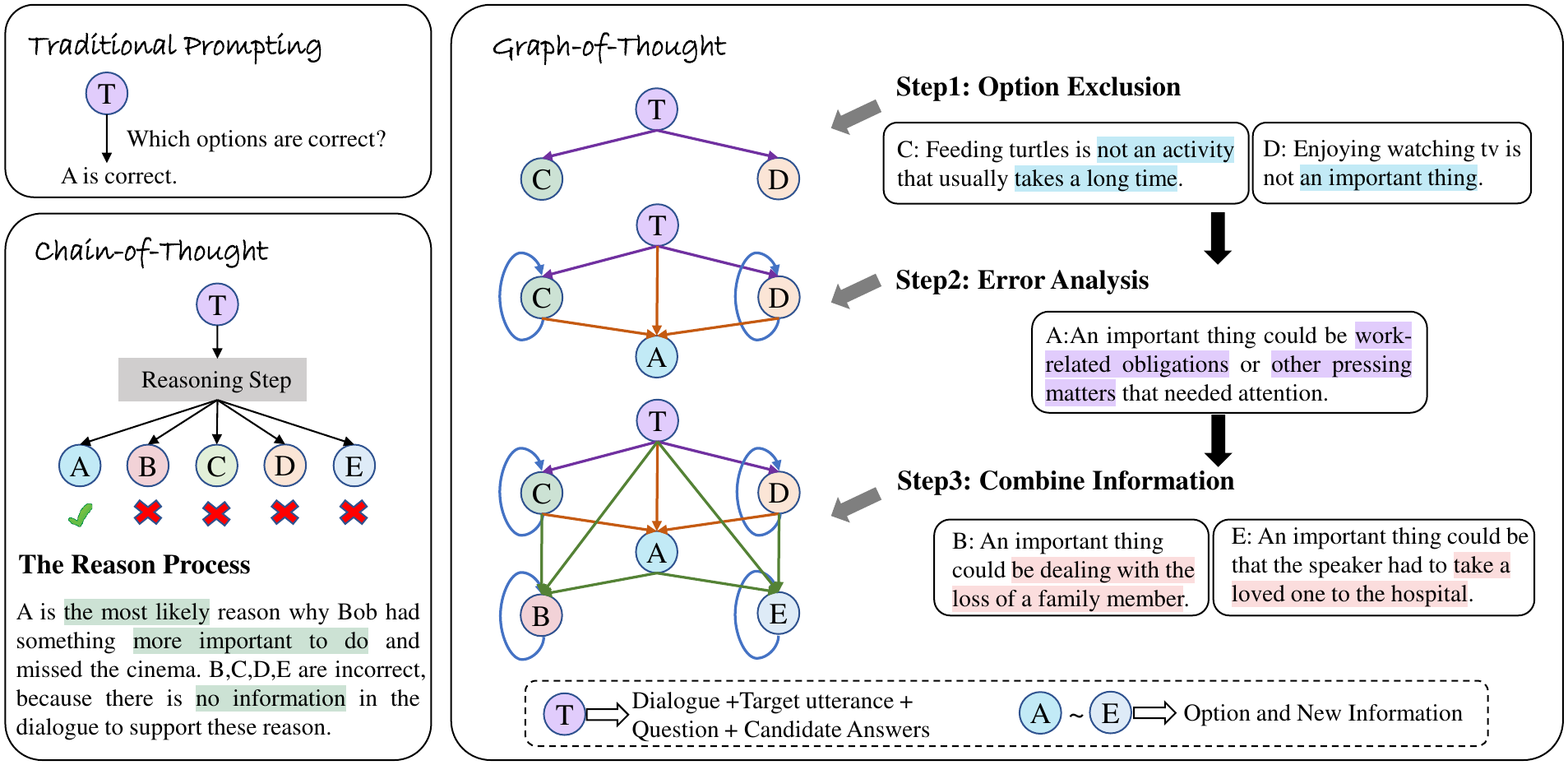}
    \caption{
    The overview of the prompt-based, CoT-based, and our ReX-GoT method.
    In our method, the purple arrows represent the first option exclusion step, \ie, leveraging the \textbf{reverse exclusion} method to effectively solve the \textbf{option saturation} challenge.
    The orange and green arrows represent the second error analysis step and the third combine information step, \ie, integrating information according to the \textbf{GoT} we design and choosing the optimal path to solve the \textbf{clue labyrinth} challenge.   
    And the blue arrows represent the updating of information at each step. 
    The highlighted text indicates the available information. }
    \label{fig:method}
    \vspace{-0.4cm}
\end{figure*}

As depicted in Figure~\ref{fig:method}, our ReX-GoT method consists of three steps. 
In the first step, the model makes an initial judgment based on the context information to exclude unreasonable options and provide the reasons for the exclusion. 
In the second step, the model conducts a detailed analysis of each option, taking into account the contextual information and the excluded options and their corresponding reasons. 
In the final step, the model synthesizes the valuable information from the first two steps for integrated reasoning and selects the optimal path for the GoT to determine the final multi-choice answer.
The specific steps are as follows.

\vpara{Step I. Option Exclusion.}
In this step, our approach involves an initial exclusion process to effectively narrow down the range of potential answers.
Subsequently, we provide the model with crucial information regarding the reasons behind the exclusion of certain options, corresponding to the ``Step$1$'' and \textit{purple arrows} in Figure~\ref{fig:method}. 
This information serves as valuable contextual input that aids the subsequent reasoning process.
Furthermore, our approach goes beyond mere exclusion by providing the model with explicit justifications for why specific options are deemed incorrect. 
By incorporating these detailed explanations into the reasoning process, we equip the model with a more comprehensive understanding of the context and enable it to engage in more informed and accurate reasoning.
Specifically, we devise the following template to consider which options are implausible and their reasons based on the given context.

\begin{mybox}\texttt
\texttt{Given the context $T$, based on common sense, which options of $u_t$ are unreasonable and why?}
\end{mybox}
This step can be formulated as:

\begin{equation}
\label{eq:2}
A_1 \leftarrow \arg\max_{\hat{\theta}} p(A_1\mid T)
\end{equation}
where $A_1$ is the text that explicitly mentions the incorrect options and their reasons, $\hat{\theta}$ means the fixed parameter of the model, as there are no gold labels in the intermediate step.
This step crucially refines the model's problem understanding, guiding subsequent reasoning by highlighting pitfalls. 
By enabling the model to recognize and comprehend excluded option reasons, it gains the ability for informed and reliable conclusions.

\vpara{Step II. Error Analysis. }
In this step, we construct a graph-of-thought to perform error analysis and comparative analysis between options based on the known information to further aid model reasoning.
Specifically, we first create a central node that represents the main stem of the problem. 
 Then, we create nodes for each answer option and their reasoning process. 
For each option, we analyze the provided information and determine if it matches the main stem of the problem. 
If it does, we mark it as a possible correct option. 
If not, we mark it as a possible incorrect option.
Next, we create a set of branch nodes for the possible correct options and analyze each branch node in more detail. 
We compare the information provided in each option with the existing information and exclude any mismatched options. 
Finally, we arrive at the correct answer by excluding the possible incorrect options and confirming that the remaining options match the provided information.

And we ask the LLM to provide detailed answers to whether each option is correct and the specific reasons, taking into account the contextual information and the unreasonable options and their corresponding reasons (just like the \textit{orange arrows} and ``Step$2$'' in Figure~\ref{fig:method}). 
The template can be viewed as follows:

\begin{mybox}\texttt
\texttt{Given $T_1 = [T, A_1]$, analysis based on the incorrect options, if the answer is $f_{t_i}$, is it reasonable and why?}
\end{mybox}

This step can be formulated as:
\begin{equation}
\label{eq:3}
A_2 \leftarrow \arg\max_{\hat{\theta}} p(A_2 \mid T_1, f_{t_i}) 
\end{equation}
where $A_2$ is the text and answer regarding whether each option is reasonable, $\hat{\theta}$ refers to the fixed parameters.
Through this step, we provide the model with the hidden information covered by the previous options and integrate the clues through the GoT for step-by-step reasoning, allowing for broader and clearer information to address the clue labyrinth challenge.

\begin{table*}[!t]
  \centering
\fontsize{9}{10}\selectfont
\setlength{\tabcolsep}{4mm}
% \resizebox{0.68\textwidth}{!}{
    \begin{tabular}{llcccccc}
    \hline
   &\multicolumn{1}{l}{\multirow{2}{*}{\textbf{Models}}}& \multicolumn{2}{c}{\textbf{CICERO}} & \multicolumn{2}{c}{\textbf{CICERO$_{v2}$}}& \multicolumn{2}{c}{\textbf{CICERO-Multi}} \\
\cmidrule(r){3-4}\cmidrule(r){5-6}\cmidrule(r){7-8}
& &F1& 	EM& 	F1 & 	EM& 	F1 & 	EM\\
 \hline
\multirow{5}{*}{$\bullet$ \textbf{\emph{SoTA baselines}}}&T5+CCID(780M) &	 81.96 &	 75.89&86.37 &	 70.14&68.03&19.95  	 \\
&T5+MCCI(780M) &	 82.64 &	76.11&87.22 &	 71.09&68.89&26.84  	 \\
&T5+TEAM(780M) &	82.73 &	 76.28&87.31&	 71.23&69.16&27.07  	 \\
&Flan-T5+TEAM(3B) &	84.46 &	 76.68&89.54&	 72.91&70.82&37.24  	 \\
&Flan-T5+TEAM(11B) &	86.62 &	 77.89&91.53&	 75.97&73.51&45.78  	 \\
\hline

\multirow{3}{*}{$\bullet$ \textbf{\emph{Prompt-based methods}}}&Flan-T5+Prompt (780M) &	82.18 &	75.74 &	86.45 &	70.21&68.78&26.71 \\
&Flan-T5+Prompt (3B) &	84.34 &	76.61 &	89.23 &	72.62&70.71&36.12\\
&Flan-T5+Prompt (11B) &	86.43 &	77.46 &	91.41 &	75.86&73.45&45.29 \\
\hline

\multirow{3}{*}{$\bullet$ \textbf{\emph{CoT-based methods}}}&Flan-T5+GoT (780M) &	84.47 &	76.72 &	87.54 &	71.37&69.83&32.94 \\
&Flan-T5+CoT (3B) &	86.52 &	77.66 &	90.63 &	74.76&72.59&40.87 \\
&Flan-T5+CoT (11B) &	87.98 &	78.53&	92.32 &	77.14&75.69&47.26 \\
\hline
\multirow{3}{*}{$\bullet$ \textbf{\emph{ReX-GoT (Ours)}}}&Flan-T5+ReX-GoT (780M) &	85.21 &	76.87 &	89.56 &	73.26&71.44&37.63 \\
&Flan-T5+ReX-GoT (3B) &	87.45 &	78.44 &	91.58 &	76.15&74.58&45.81 \\
&Flan-T5+ReX-GoT (11B) &	\textbf{89.52} &	\textbf{80.63} &	\textbf{93.87} &	\textbf{78.46}&\textbf{78.51}&\textbf{53.08} \\
\hline

    \end{tabular}%
    % }
    \caption{Comparison of our method with baselines on CICERO, CICERO$_{v2}$ and CICERO-Multi datasets. CICERO-Multi datasets is a subdataset within the CICERO dataset that only contained questions with multiple correct answer choices.}
  \label{tab:overall}%
  \vspace{-4mm}
\end{table*}%

\vpara{Step III. Combine Information.}
In this step, as the ``Step$3$'' and \textit{green arrows} in Figure~\ref{fig:method} show,
we leverage the valuable insights gathered from the preceding two steps and employ the GoT to further advance our reasoning process. 
Specifically, in inference steps I and II, we set the LLM decoder to generate multiple answers as different paths through the GoT, each of which give a different prediction for each option.
The final multi-choice answer is determined by selecting the optimal path through a voting mechanism.
With the aid of GoT, we delve into the intricate nuances of the more complex and challenging options, persisting until a comprehensive evaluation of all options are achieved. 
This diligent examination ultimately culminates in the determination of the final multi-choice answer $\hat{y}$.
The template can be viewed as follows:

\begin{mybox}\texttt
\texttt{Given $T_2 = [T_1, A_2]$, analysis based on the previous steps, which options of $u_t$ are reasonable?}
\end{mybox}

This step can be formulated as:
\begin{equation}
\label{eq:4}
\hat{y} \leftarrow \arg\max_{\theta} p(y \mid T_2)
\end{equation}
where $\theta$ can be fine-tuned during training with the gold task annotations.

Each step considers new information and validates the previous reasoning to arrive at the correct answer.  
By utilizing GOT, we visually represent the reasoning process and enable our method to integrate sophisticated reasoning clues.
In conclusion, our ReX-GoT takes into account the subtle details and intricate dependencies in the given context and options. 
It combines multiple predictive information and guides the model to reason step-by-step through the GoT in a reverse exclusion manner. 
This approach effectively addresses the challenges of option saturation and clue labyrinth in DC-MCQ.

%% file: Section/4-experiments.tex
\section{Experiments}

\subsection{Implementation Details}
\textbf{Datasets.} We assess the efficacy of models on two benchmark datasets, CICERO~\cite{ghosal2022cicero} and CICERO$_{v2}$~\cite{shen2022multiview}. CICERO is a binary dialogue dataset featuring five types of dialogue-level inferences: causality, consequence, premise, motivation, and emotional reaction. 
The dataset comprises 53,105 inferences from 5,672 dialogues. 
CICERO$_{v2}$ is built upon the original CICERO dataset, where only 15\% of the inferences in CICERO are multi-choice, whereas all 8,351 inferences in CICERO$_{v2}$ are multi-choice. The dataset consists of 2,379 dialogues.

\noindent \textbf{Evaluation Metrics.}
We use macro-F1 and Exact Match as evaluation metrics for our models. Macro-F1 considers precision and recall across multiple classes and provides an average score. Exact Match measures the percentage of correct predictions that exactly match the expected answers. 
All our scores are the average over 5 runnings with random seeds.

\noindent \textbf{Settings.} 
Due to the outstanding performance of Flan-T5, a encoder-decoder style language model, we utilize it as the backbone LLM for our method. 
We also test with GPT3.5. 
We use four versions of Flan-T5: 250M (base), 780M (large), 3B (xl), and 11B (11B). 
Our experiments are conducted using NVIDIA A100 GPUs. 

\noindent \textbf{Baseline Systems.}
We compare our method with the state-of-the-art (SoTA) baselines, including
\begin{itemize}

\item \textbf{CCID:} \citet{ghosal2022cicero} computed a match by comparing each generated answer to a candidate selection. 

\item \textbf{MCCI:} \citet{shen2022multiview} proposed a pre-trained transformer DIALeCT for dialogue commonsense inference.
\item \textbf{TEAM:} \citet{ghosal2022two} simply refactored the multi-choice question answering task into a series of binary classifications.
\end{itemize}

\subsection{Overall Results}

We first comprehensively evaluate our ReX-GoT's superiority in dialogue commonsense inference using F1 and EM metrics. 
We compare against SoTA baselines (CCID, MCCI, TEAM), Prompt-based, and CoT-based methods across CICERO, CICERO$_{v2}$, and CICERO-Multi datasets.
Table~\ref{tab:overall} highlights ReX-GoT's advantage over SoTA baselines. 
Flan-T5-large exhibits notable improvement with ReX-GoT. 
Further, with an 11B-parameter LLM, ReX-GoT outperforms the best baseline TEAM, \eg, on CICERO, by 2.9\% in F1 score and 2.74\% in EM score.
On CICERO$_{v2}$, ReX-GoT surpasses the SoTA baseline TEAM by 2.34\% in F1 score and 2.49\% in EM score. 
Moreover, our ReX-GoT exhibits a remarkable enhancement compared to vanilla prompting and CoT methods, particularly on the CICERO dataset with multiple correct answer options, where the EM scores of our model improve by 5.82\% and 7.79\%, respectively.
These findings suggest that our ReX-GoT can make use of the hidden information between options to enhance reasoning and make answers more explanatory than vanilla prompting and CoT methods. 
Notably, our ReX-GoT effectively addresses the option saturation and clue labyrinth challenges in dialogue commonsense inference.

%\vspace{-2mm}
\subsection{Results on Zero-shot Inference}
%\vspace{-1mm}

\begin{table}[!t]
  \centering
\fontsize{9}{11.5}\selectfont
\setlength{\tabcolsep}{2mm}
\resizebox{0.98\columnwidth}{!}{
\begin{tabular}{lcccc}
\hline

\multicolumn{1}{c}{\multirow{2}{*}{\textbf{}}}& \multicolumn{2}{c}{\textbf{CICERO}} & \multicolumn{2}{c}{\textbf{CICERO$_{v2}$}} \\
\cmidrule(r){2-3}\cmidrule(r){4-5}
& F1& 	EM& 	F1 & 	EM\\
 \hline
\multicolumn{5}{l}{$\bullet$ \textbf{\emph{SoTA baselines}}} \\
Flan-T5+TEAM(3B) &	 47.95 &	 42.11&48.79 &	 40.64  	 \\
Flan-T5+TEAM(11B) &	 51.21 &	 45.53&51.68 &	 42.75  	 \\

\hline

\multicolumn{5}{l}{$\bullet$ \textbf{\emph{Prompt-based methods}}}\\

Flan-T5+Prompt (3B) &	48.34 &	42.53 &	49.28 &	41.67 \\
Flan-T5+Prompt (11B) &	51.67 &	45.82 &	52.34 &	43.29 \\
\hline

\multicolumn{5}{l}{$\bullet$ \textbf{\emph{CoT-based methods}}}\\
Flan-T5+CoT (3B) &	54.48 &	47.64 &	55.69 &	44.72 \\
Flan-T5+CoT (11B) &	58.83 &	49.77 &	60.22 &	47.26 \\
\hline
\multicolumn{5}{l}{$\bullet$ \textbf{\emph{ReX-GoT (Ours)}}}\\

Flan-T5+ReX-GoT (3B) &	63.59 &	52.84 &	64.38 &	49.64 \\
Flan-T5+ReX-GoT (11B) &	67.73 &	55.39 &	69.35 &	53.33 \\
GPT3.5+ReX-GoT  &	86.04 &	77.17 &	91.12 &	75.73 \\
\hline

\end{tabular}%
}
\caption{
Experimental results on zero-shot setting.}
  \vspace{-4mm}
  \label{tab:zero}%
\end{table}%

We conduct a comprehensive comparison of our proposed ReX-GoT with SoTA approaches, Prompt-based, and CoT-based methods under zero-shot conditions. The results in Table~\ref{tab:zero} demonstrate our method's supremacy across all metrics.
Prompt-based and CoT-based techniques exhibit substantial improvements over the current SoTA baseline. 
However, our ReX-GoT approach stands out with even more substantial advancements in dialogue commonsense inference. 
As an example, on the CICERO$_{v2}$ dataset, when using Flan-T5-11B, our ReX-GoT approach demonstrates a remarkable improvement in F1 score of 17.67\% over the best-performing baseline TEAM. 
Our ReX-GoT approach outperforms the prompt-based approach by a margin of 17.01\% in F1 score and the CoT-based approach by a margin of 9.13\% in F1 score. 
Remarkably, when integrated into an ultra-large LLM like GPT3.5-175B, ReX-GoT achieves remarkable improvements, enhancing the SoTA's F1 score by 34.83\% on CICERO and 39.44\% on CICERO$_{v2}$.
These results highlight the effectiveness of our ReX-GoT approach in improving the performance of large language models on dialogue commonsense inference.

\subsection{Ablation Study}

\begin{table}[!t]
  \centering
\fontsize{9}{11}\selectfont
\setlength{\tabcolsep}{0.7mm}
\resizebox{1\columnwidth}{!}{
    \begin{tabular}{lcccc}
    \hline
   \multicolumn{1}{c}{\multirow{2}{*}{\textbf{}}}& \multicolumn{2}{c}{\textbf{CICERO}} & \multicolumn{2}{c}{\textbf{CICERO-Multi}} \\
\cmidrule(r){2-3}\cmidrule(r){4-5}
& F1& 	EM& 	F1 & 	EM\\
 \hline

ReX-GoT &	89.52 &	80.63 &78.51&53.08 \\
w/o CI&88.39(-1.13)&79.41(-1.22)&77.71(-0.8)
&51.66(-1.42) \\
w/o ReX&88.02(-1.50)&78.97(-1.66)&77.54(-0.97)&51.39(-1.69) \\
w/o GoT&87.25(-2.27)&78.17(-2.46)&76.33(-2.18)&49.43(-3.65)\\

\hline

    \end{tabular}%
    }
    \caption{
    Ablation results on \text{DC-MCQ} task. CI means the combined information step, ReX means the reverse exclusion step.  
    In the brackets are the drops than ReX-GoT.
    }
    %\vspace{-4mm}
  \label{tab:ablation}%
\end{table}%

We conduct ablation experiments to evaluate the contribution of each component in our model. 
As depicted in Table \ref{tab:ablation}, no variant matches the full model's performance, highlighting the indispensability of each component.
Specifically, the F1 score drops most severely when the graph-of-thought are not used, which suggests that guiding the model to reason step-by-step and considering hidden information among options is crucial.  
To verify the necessity and effectiveness of exclusion, we remove the exclusion step, and the sharp drop in the results demonstrates its unignorable effect on dialogue commonsense inference. 
This finding suggests that combining exclusion with forward reasoning is essential for improving the performance of our model.
In addition, removing combine information step leads to a marked drop in performance, indicating the importance of combining intricate clues in our ReX-GoT method.

\subsection{Analyses and Discussions}

To further investigate the effectiveness of ReX-GoT, we conduct in-depth analyses to answer the following questions, with the aim to reveal how our proposed methods advance.

\paragraph{How multi-choice inference affect model performance?}

\pgfplotsset{compat=newest}

\begin{figure}[!t]
\centering
\begin{tikzpicture}
[font=\small]
%[font=\scriptsize]
\begin{axis}[
    ybar,
    height=0.625\columnwidth,
    width=0.9\columnwidth,
    enlargelimits=0.13,
    legend style={at={(0.3,0.86)},
      anchor=north,legend columns=-1},
    ylabel={EM Score (\%)},
    symbolic x coords={TEAM,Prompt,CoT,ReX-GoT,GPT3.5},
    xtick=data,
    ymin= 10, ymax= 80,
    ybar=3pt,
    bar width=8pt,
    nodes near coords,
    nodes near coords align={vertical},
    ]

\addplot coordinates {(TEAM,45.53) (Prompt,45.82) (CoT,49.77) (ReX-GoT,55.39) (GPT3.5,77.17)};
\addplot coordinates {(TEAM,22.79) (Prompt,23.67) (CoT,29.25) (ReX-GoT,37.48)(GPT3.5,51.39)};

\legend{All,Multi}
\end{axis}
\end{tikzpicture}
    %\vspace{-2mm}
\caption{
Comparison with different models on dialogue
commonsense inference. ``All'' and ``Multi'' mean that the results are calculated on the complete CICERO dataset and a subset of CICERO containing only multiple correct options.
}
\label{fig:multi}
  \vspace{-5mm}
\end{figure}
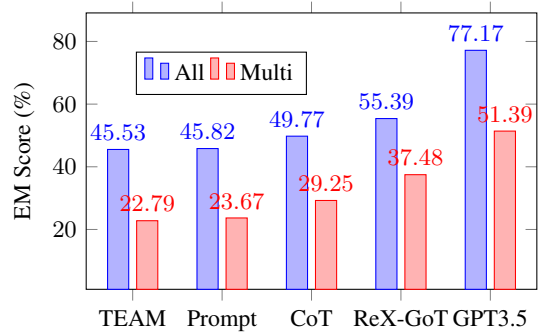

We are curious about the impact of multi-choice inference on model performance in an unsupervised setting.
In Figure~\ref{fig:multi}, we compare our model with the best baseline models on the Multi and All datasets, as well as models based on vanilla prompting and CoT. 
Our findings show that our model consistently outperforms these models in dialogue commonsense inference, regardless of whether the single-choice or multi-choice questions.
Furthermore, the performance gaps are further enlarged when considering multi-choice inference, indicating the effectiveness of our method for this task. 
Overall, our results highlight the potential of our method for multi-choice inference in unsupervised conditions. 
By effectively integrating available clues from answer options, our approach surpass existing baseline models, even in challenging scenarios with multiple correct answer options.

\paragraph{How the number of correct options affect model performance?}

\begin{figure}[!t]   
  \centering
     \subcaptionbox{F1 Score}{\includegraphics[width=0.23\textwidth]{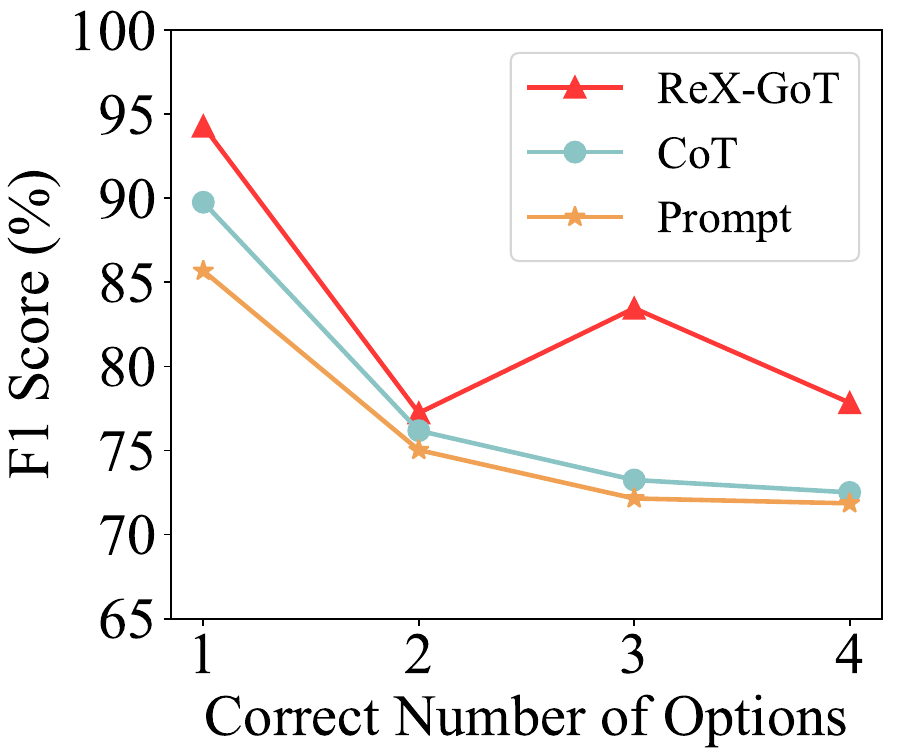}}
         \subcaptionbox{EM Score}{\includegraphics[width=0.23\textwidth]{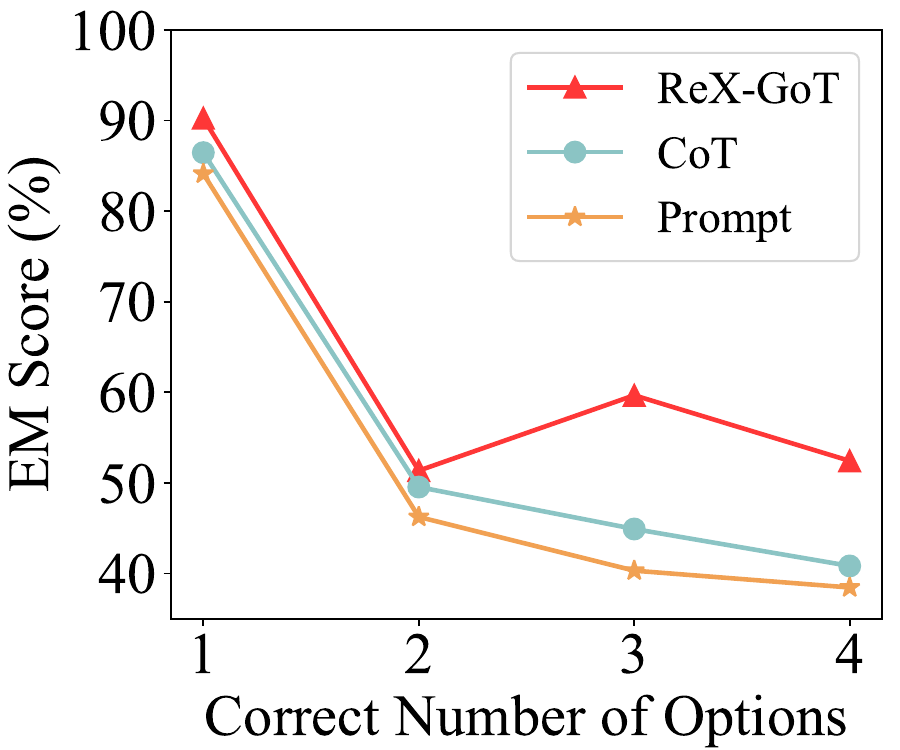}}
      \caption{Influence of correct number of options. 
      }
      % \vspace{-5mm}
       \label{fig:option number}
\end{figure}

We investigate the effect of the correct number of options on our model's performance in dialogue commonsense inference.
As shown in Figure~\ref{fig:option number}, we observe that the model's performance varies with the number of correct options. 
Our ReX-GoT method performs worst on questions with two correct options, followed by questions with four, three, and performs best on questions with one correct option. 
On the other hand, vanilla prompting and CoT methods show a decline in performance as the number of correct options increases.
ReX-GoT effectively utilizes option information, capturing the relationship between options and context to differentiate between correct and incorrect options. This advantage is particularly prominent in questions with multiple correct options, where option information plays a crucial role. 
In contrast, vanilla methods rely only on context, neglects the integration of hidden clues and underutilizes the additional information in the options.

\paragraph{What are the advantages of ReX-GoT over forward reasoning and backward exclusion?}

\begin{figure}[!t]   
  \centering
     \subcaptionbox{F1 Score}{\includegraphics[width=0.23\textwidth]{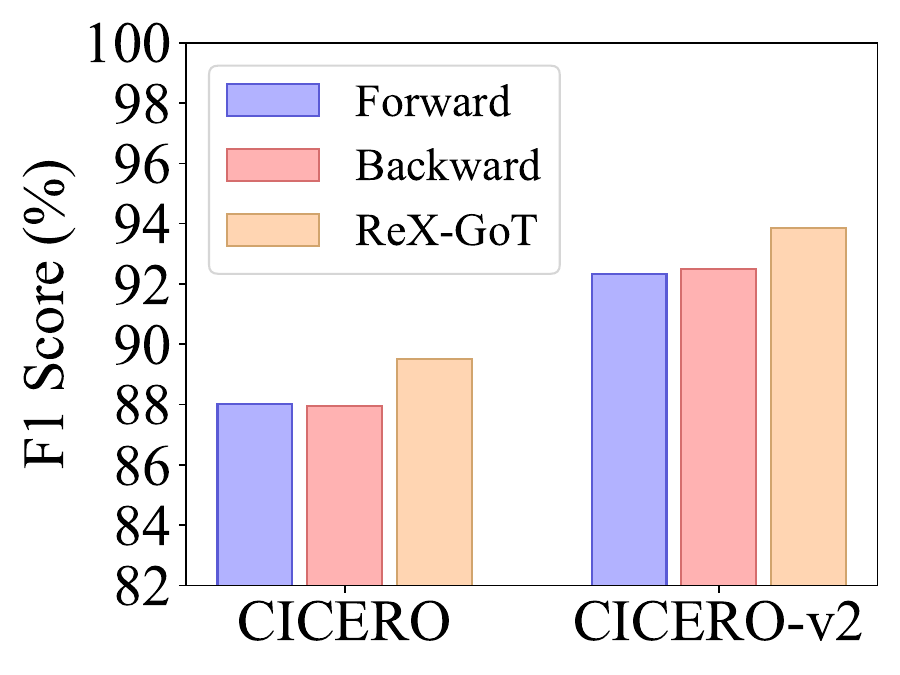}}
         \subcaptionbox{EM Score}{\includegraphics[width=0.2213\textwidth]{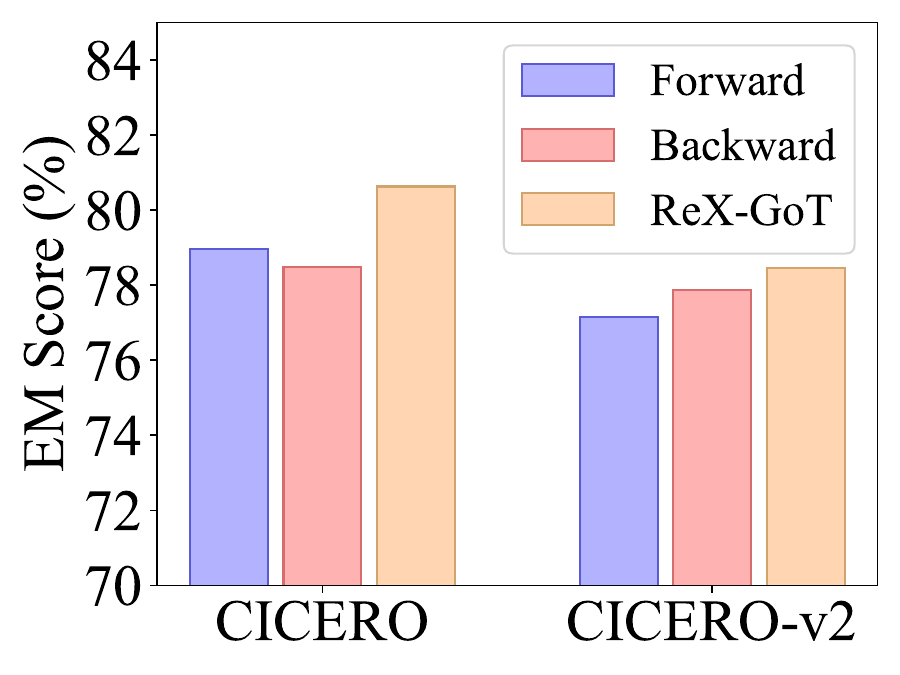}}
      \caption{Influence of different prompting. 
      }
      \vspace{-0.4cm}
       \label{fig:prompt}
        
\end{figure}

We conduct experiments to compare our ReX-GoT approach with forward reasoning and backward exclusion. The results in Figure~\ref{fig:prompt} show that ReX-GoT outperforms the two single methods on both datasets. 
Forward reasoning involves selecting the most plausible option at each step until no correct option is left.
Backward exclusion, on the other hand, involves selecting the most incorrect option at each step until no incorrect options remain.
Interestingly, the performance of the individual methods reverses between the datasets. 
In the CICERO dataset, forward reasoning is superior, while in the CICERO${v_2}$ dataset, backward exclusion performs better. 
This reversal is attributed to the majority of single-choice questions in the CICERO dataset, where inadequate exclusion during backward exclusion leads to decreased performance. 
Conversely, the CICERO${v_2}$ dataset consists exclusively of multi-choice questions, making forward reasoning more challenging and resulting in poorer performance compared to backward exclusion and ReX-GoT. 
These findings further support the necessity of designing ReX-GoT for multi-choice task as it  effectively combines the two single approaches and integrates valuable clues to address challenges and improve overall performance.

\paragraph{How LLMs scales affect model performance?}

\begin{figure}[!t]   
  \centering
      \subcaptionbox{F1 Score}{\includegraphics[width=0.23\textwidth]{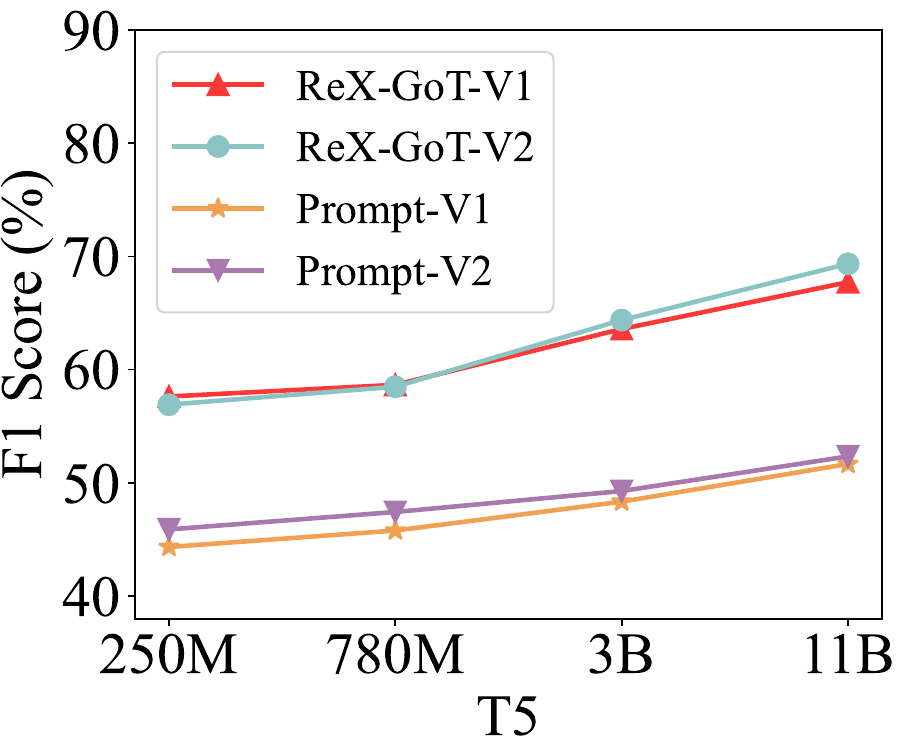}}
     \subcaptionbox{EM Score}{\includegraphics[width=0.23\textwidth]{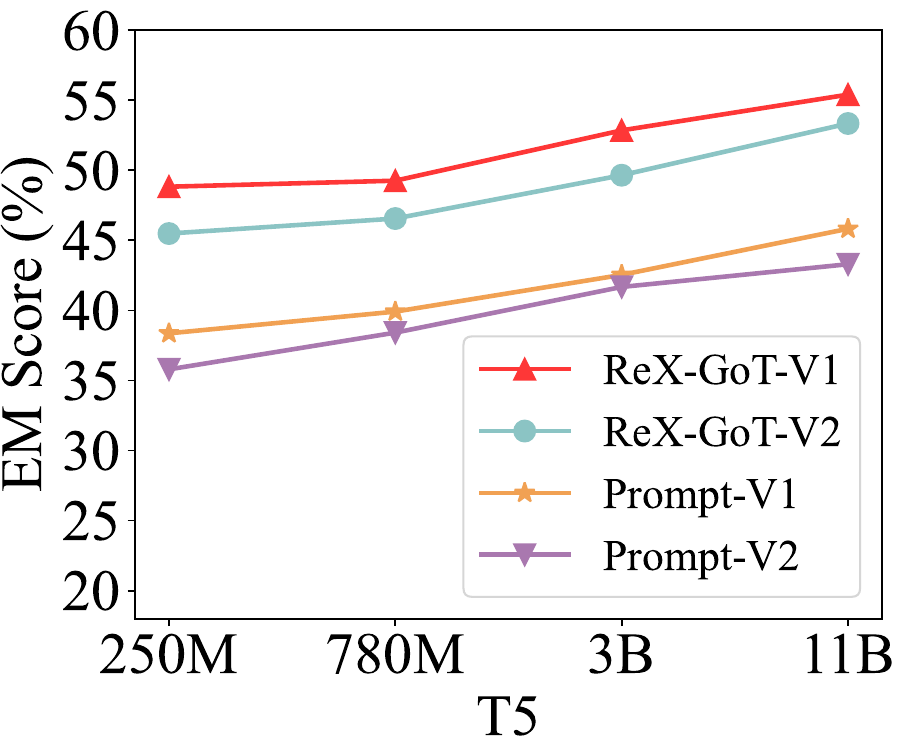}}
      \caption{Influences of LLM scales.}
      \vspace{-4mm}
       \label{fig:scale}
\end{figure}

We study the effect of different LLMs scales and provide the experimental results in Figure~\ref{fig:scale}. 
We observe that both prompt-based and ReX-GoT methods show
notable performance enhancement as the model size increases, particularly from Flan-T5-3B to Flan-T5-11B, where our ReX-GoT approach achieves an increase of 4.97\% in F1 score and 3.69\% in EM score on the CICERO$_{v2}$ dataset.
Our results are consistent with existing findings about the effectiveness of CoT prompts, indicating that larger LLMs can bring remarkable improvements. 
This is because larger LLMs have stronger abilities to capture and model complex patterns and relationships in the data.
Overall, our results emphasize the importance of considering LLMs size when designing models for dialogue commonsense inference.
These findings demonstrate the potential of using LLMs for this task. 

%% file: Section/5-conclusion.tex
\section{Conclusion}

In this paper, we address the pressing option saturation and clue labyrinth challenges in the Dialogue Commonsense Multi-choice Question Answering task. 
We propose ReX-GoT, a novel three-step Reverse Exclusion Graph-of-Thought framework including Option Exclusion, Error Analysis, and Combine Information to mimic human reasoning.
Through the gradual exclusion of irrelevant options and the incorporation of human-like reasoning, the final answer is obtained by constructing a GoT and selecting its optimal path.
Our extensive experimental results on CICERO and CICERO$_{v2}$ datasets demonstrate that our scheme achieves SoTA performance on both single-choice and multi-choice dialogue commonsense inference.

%% file: main.bbl
\begin{thebibliography}{35}
\providecommand{\natexlab}[1]{#1}

\bibitem[{Arabshahi et~al.(2021)Arabshahi, Lee, Gawarecki, Mazaitis, Azaria,
  and Mitchell}]{DBLP:conf/aaai/ArabshahiLGMAM21}
Arabshahi, F.; Lee, J.; Gawarecki, M.; Mazaitis, K.; Azaria, A.; and Mitchell,
  T.~M. 2021.
\newblock Conversational Neuro-Symbolic Commonsense Reasoning.
\newblock In \emph{Proceedings of the 35th AAAI Conference on Artificial
  Intelligence (AAAI'21)}, 4902--4911.

\bibitem[{Bosselut, Bras, and Choi(2021)}]{DBLP:conf/aaai/BosselutBC21}
Bosselut, A.; Bras, R.~L.; and Choi, Y. 2021.
\newblock Dynamic Neuro-Symbolic Knowledge Graph Construction for Zero-shot
  Commonsense Question Answering.
\newblock In \emph{Proceedings of the 35th AAAI Conference on Artificial
  Intelligence (AAAI'21)}, 4923--4931.

\bibitem[{Brown et~al.(2020)Brown, Mann, Ryder, Subbiah, Kaplan, Dhariwal,
  Neelakantan, Shyam, Sastry, Askell, Agarwal, Herbert{-}Voss, Krueger,
  Henighan, Child, Ramesh, Ziegler, Wu, Winter, Hesse, Chen, Sigler, Litwin,
  Gray, Chess, Clark, Berner, McCandlish, Radford, Sutskever, and
  Amodei}]{DBLP:conf/nips/BrownMRSKDNSSAA20}
Brown, T.~B.; Mann, B.; Ryder, N.; Subbiah, M.; Kaplan, J.; Dhariwal, P.;
  Neelakantan, A.; Shyam, P.; Sastry, G.; Askell, A.; Agarwal, S.;
  Herbert{-}Voss, A.; Krueger, G.; Henighan, T.; Child, R.; Ramesh, A.;
  Ziegler, D.~M.; Wu, J.; Winter, C.; Hesse, C.; Chen, M.; Sigler, E.; Litwin,
  M.; Gray, S.; Chess, B.; Clark, J.; Berner, C.; McCandlish, S.; Radford, A.;
  Sutskever, I.; and Amodei, D. 2020.
\newblock Language Models are Few-Shot Learners.
\newblock In Larochelle, H.; Ranzato, M.; Hadsell, R.; Balcan, M.; and Lin, H.,
  eds., \emph{Proceedings of the 34th International Conference on Neural
  Information Processing Systems (NeurIPS'20)}.

\bibitem[{Chen et~al.(2023)Chen, Xu, Yan, Zhang, Huang, Si, and
  Zhang}]{DBLP:conf/acl/ChenXYZHSZ23}
Chen, Q.; Xu, G.; Yan, M.; Zhang, J.; Huang, F.; Si, L.; and Zhang, Y. 2023.
\newblock Distinguish Before Answer: Generating Contrastive Explanation as
  Knowledge for Commonsense Question Answering.
\newblock In Rogers, A.; Boyd{-}Graber, J.~L.; and Okazaki, N., eds.,
  \emph{Findings of the 61st Annual Meeting of the Association for
  Computational Linguistics (ACL'23)}, 13207--13224. Association for
  Computational Linguistics.

\bibitem[{Deng et~al.(2023)Deng, Liu, Li, Wang, Zhang, Li, Wang, Zhang, and
  Liu}]{DBLP:journals/corr/abs-2307-08715}
Deng, G.; Liu, Y.; Li, Y.; Wang, K.; Zhang, Y.; Li, Z.; Wang, H.; Zhang, T.;
  and Liu, Y. 2023.
\newblock Jailbreaker: Automated Jailbreak Across Multiple Large Language Model
  Chatbots.
\newblock \emph{CoRR}, abs/2307.08715.

\bibitem[{Dou and Peng(2022)}]{DBLP:conf/aaai/DouP22}
Dou, Z.; and Peng, N. 2022.
\newblock Zero-Shot Commonsense Question Answering with Cloze Translation and
  Consistency Optimization.
\newblock In \emph{Proceedings of the 36th AAAI Conference on Artificial
  Intelligence (AAAI'22)}, 10572--10580. {AAAI} Press.

\bibitem[{Fei et~al.(2023)Fei, Li, Liu, Bing, Li, and Chua}]{fei2023reasoning}
Fei, H.; Li, B.; Liu, Q.; Bing, L.; Li, F.; and Chua, T.-S. 2023.
\newblock Reasoning Implicit Sentiment with Chain-of-Thought Prompting.
\newblock \emph{arXiv preprint arXiv:2305.11255}.

\bibitem[{Ghosal et~al.(2021)Ghosal, Hong, Shen, Majumder, Mihalcea, and
  Poria}]{DBLP:conf/sigdial/GhosalHSMMP21}
Ghosal, D.; Hong, P.; Shen, S.; Majumder, N.; Mihalcea, R.; and Poria, S. 2021.
\newblock {CIDER:} Commonsense Inference for Dialogue Explanation and
  Reasoning.
\newblock In Li, H.; Levow, G.; Yu, Z.; Gupta, C.; Sisman, B.; Cai, S.;
  Vandyke, D.; Dethlefs, N.; Wu, Y.; and Li, J.~J., eds., \emph{Proceedings of
  the 22nd Annual Meeting of the Special Interest Group on Discourse and
  Dialogue (SIGdial'21)}, 301--313. Association for Computational Linguistics.

\bibitem[{Ghosal et~al.(2022{\natexlab{a}})Ghosal, Majumder, Mihalcea, and
  Poria}]{ghosal2022two}
Ghosal, D.; Majumder, N.; Mihalcea, R.; and Poria, S. 2022{\natexlab{a}}.
\newblock Two is Better than Many Binary? Classification as an Effective
  Approach to Multi-Choice Question Answering.
\newblock \emph{arXiv preprint arXiv:2210.16495}.

\bibitem[{Ghosal et~al.(2022{\natexlab{b}})Ghosal, Shen, Majumder, Mihalcea,
  and Poria}]{ghosal2022cicero}
Ghosal, D.; Shen, S.; Majumder, N.; Mihalcea, R.; and Poria, S.
  2022{\natexlab{b}}.
\newblock CICERO: A Dataset for Contextualized Commonsense Inference in
  Dialogues.
\newblock In \emph{Proceedings of the 60th Annual Meeting of the Association
  for Computational Linguistics (ACL'22)}, 5010--5028.

\bibitem[{Imani, Du, and Shrivastava(2023)}]{DBLP:conf/acl/ImaniD023}
Imani, S.; Du, L.; and Shrivastava, H. 2023.
\newblock MathPrompter: Mathematical Reasoning using Large Language Models.
\newblock In Sitaram, S.; Klebanov, B.~B.; and Williams, J.~D., eds.,
  \emph{Proceedings of the The 61st Annual Meeting of the Association for
  Computational Linguistics (ACL'23)}, 37--42. Association for Computational
  Linguistics.

\bibitem[{Jin and Lu(2023)}]{DBLP:conf/acl/JinL23}
Jin, Z.; and Lu, W. 2023.
\newblock Tab-CoT: Zero-shot Tabular Chain of Thought.
\newblock In Rogers, A.; Boyd{-}Graber, J.~L.; and Okazaki, N., eds.,
  \emph{Findings of the 61st Annual Meeting of the Association for
  Computational Linguistics (ACL'23)}, 10259--10277. Association for
  Computational Linguistics.

\bibitem[{Ju et~al.(2021)Ju, Zhang, Tian, Liu, Cao, Zhao, Li, and
  Zhao}]{ju2021enhancing}
Ju, Y.; Zhang, Y.; Tian, Z.; Liu, K.; Cao, X.; Zhao, W.; Li, J.; and Zhao, J.
  2021.
\newblock Enhancing multiple-choice machine reading comprehension by punishing
  illogical interpretations.
\newblock In \emph{Proceedings of the Conference on Empirical Methods in
  Natural Language Processing (EMNLP'21)}, 3641--3652.

\bibitem[{Kuo and Chen(2023)}]{DBLP:conf/acl/KuoC23}
Kuo, H.; and Chen, Y. 2023.
\newblock Zero-Shot Prompting for Implicit Intent Prediction and Recommendation
  with Commonsense Reasoning.
\newblock In Rogers, A.; Boyd{-}Graber, J.~L.; and Okazaki, N., eds.,
  \emph{Findings of the 61st Annual Meeting of the Association for
  Computational Linguistics (ACL'23)}, 249--258. Association for Computational
  Linguistics.

\bibitem[{Liu et~al.(2022)Liu, Liu, Lu, Welleck, West, Le~Bras, Choi, and
  Hajishirzi}]{DBLP:conf/acl/0010LLWWBCH22}
Liu, J.; Liu, A.; Lu, X.; Welleck, S.; West, P.; Le~Bras, R.; Choi, Y.; and
  Hajishirzi, H. 2022.
\newblock Generated Knowledge Prompting for Commonsense Reasoning.
\newblock In \emph{Proceedings of the 60th Annual Meeting of the Association
  for Computational Linguistics (ACL'22)}, 3154--3169.

\bibitem[{Ma et~al.(2021)Ma, Ilievski, Francis, Bisk, Nyberg, and
  Oltramari}]{DBLP:conf/aaai/MaIFBNO21}
Ma, K.; Ilievski, F.; Francis, J.; Bisk, Y.; Nyberg, E.; and Oltramari, A.
  2021.
\newblock Knowledge-driven Data Construction for Zero-shot Evaluation in
  Commonsense Question Answering.
\newblock In \emph{Proceedings of the 35th AAAI Conference on Artificial
  Intelligence (AAAI'21)}, 13507--13515.

\bibitem[{Ma et~al.(2023)Ma, Yu, Li, and Li}]{DBLP:conf/aaai/MaYLL23}
Ma, Z.; Yu, Z.; Li, J.; and Li, G. 2023.
\newblock HybridPrompt: Bridging Language Models and Human Priors in Prompt
  Tuning for Visual Question Answering.
\newblock In Williams, B.; Chen, Y.; and Neville, J., eds., \emph{Proceedings
  of the 37th AAAI Conference on Artificial Intelligence (AAAI'23)},
  13371--13379.

\bibitem[{Ouyang et~al.(2022)Ouyang, Wu, Jiang, Almeida, Wainwright, Mishkin,
  Zhang, Agarwal, Slama, Ray, Schulman, Hilton, Kelton, Miller, Simens, Askell,
  Welinder, Christiano, Leike, and Lowe}]{ouyang2022training}
Ouyang, L.; Wu, J.; Jiang, X.; Almeida, D.; Wainwright, C.~L.; Mishkin, P.;
  Zhang, C.; Agarwal, S.; Slama, K.; Ray, A.; Schulman, J.; Hilton, J.; Kelton,
  F.; Miller, L.; Simens, M.; Askell, A.; Welinder, P.; Christiano, P.~F.;
  Leike, J.; and Lowe, R. 2022.
\newblock Training language models to follow instructions with human feedback.
\newblock In \emph{Proceedings of the 36th International Conference on Neural
  Information Processing Systems (NeurIPS'22)}, 27730--27744.

\bibitem[{Paranjape et~al.(2021)Paranjape, Michael, Ghazvininejad, Hajishirzi,
  and Zettlemoyer}]{DBLP:conf/acl/ParanjapeMGHZ21}
Paranjape, B.; Michael, J.; Ghazvininejad, M.; Hajishirzi, H.; and Zettlemoyer,
  L. 2021.
\newblock Prompting Contrastive Explanations for Commonsense Reasoning Tasks.
\newblock In Zong, C.; Xia, F.; Li, W.; and Navigli, R., eds., \emph{Findings
  of the Association for Computational Linguistics: {ACL-IJCNLP} 2021, Online
  Event, August 1-6, 2021}, volume {ACL-IJCNLP} 2021 of \emph{Findings of
  {ACL}}, 4179--4192. Association for Computational Linguistics.

\bibitem[{Qin et~al.(2021)Qin, Gupta, Upadhyay, He, Choi, and
  Faruqui}]{DBLP:conf/acl/QinGUHCF20}
Qin, L.; Gupta, A.; Upadhyay, S.; He, L.; Choi, Y.; and Faruqui, M. 2021.
\newblock {TIMEDIAL:} Temporal Commonsense Reasoning in Dialog.
\newblock In Zong, C.; Xia, F.; Li, W.; and Navigli, R., eds.,
  \emph{Proceedings of the 59th Annual Meeting of the Association for
  Computational Linguistics and the 11th International Joint Conference on
  Natural Language Processing, {ACL/IJCNLP} 2021, (Volume 1: Long Papers),
  Virtual Event, August 1-6, 2021}, 7066--7076. Association for Computational
  Linguistics.

\bibitem[{Richardson and Heck(2023)}]{DBLP:journals/corr/abs-2302-07926}
Richardson, C.; and Heck, L. 2023.
\newblock Commonsense Reasoning for Conversational {AI:} {A} Survey of the
  State of the Art.
\newblock \emph{CoRR}, abs/2302.07926.

\bibitem[{Shen et~al.(2022)Shen, Ghosal, Majumder, Lim, Mihalcea, and
  Poria}]{shen2022multiview}
Shen, S.; Ghosal, D.; Majumder, N.; Lim, H.; Mihalcea, R.; and Poria, S. 2022.
\newblock Multiview contextual commonsense inference: A new dataset and task.
\newblock \emph{arXiv preprint arXiv:2210.02890}.

\bibitem[{Trivedi et~al.(2023)Trivedi, Balasubramanian, Khot, and
  Sabharwal}]{DBLP:conf/acl/TrivediBKS23}
Trivedi, H.; Balasubramanian, N.; Khot, T.; and Sabharwal, A. 2023.
\newblock Interleaving Retrieval with Chain-of-Thought Reasoning for
  Knowledge-Intensive Multi-Step Questions.
\newblock In Rogers, A.; Boyd{-}Graber, J.~L.; and Okazaki, N., eds.,
  \emph{Proceedings of the 61st Annual Meeting of the Association for
  Computational Linguistics (ACL'23)}, 10014--10037. Association for
  Computational Linguistics.

\bibitem[{Wang et~al.(2023{\natexlab{a}})Wang, Xu, Lan, Hu, Lan, Lee, and
  Lim}]{DBLP:conf/acl/WangXLHLLL23}
Wang, L.; Xu, W.; Lan, Y.; Hu, Z.; Lan, Y.; Lee, R.~K.; and Lim, E.
  2023{\natexlab{a}}.
\newblock Plan-and-Solve Prompting: Improving Zero-Shot Chain-of-Thought
  Reasoning by Large Language Models.
\newblock In Rogers, A.; Boyd{-}Graber, J.~L.; and Okazaki, N., eds.,
  \emph{Proceedings of the 61st Annual Meeting of the Association for
  Computational Linguistics (ACL'23)}, 2609--2634. Association for
  Computational Linguistics.

\bibitem[{Wang et~al.(2023{\natexlab{b}})Wang, Wang, Li, Gao, Yin, and
  Ren}]{DBLP:conf/acl/WangWLGYR23}
Wang, P.; Wang, Z.; Li, Z.; Gao, Y.; Yin, B.; and Ren, X. 2023{\natexlab{b}}.
\newblock {SCOTT:} Self-Consistent Chain-of-Thought Distillation.
\newblock In Rogers, A.; Boyd{-}Graber, J.~L.; and Okazaki, N., eds.,
  \emph{Proceedings of the 61st Annual Meeting of the Association for
  Computational Linguistics (ACL'23)}, 5546--5558. Association for
  Computational Linguistics.

\bibitem[{Wang et~al.(2018)Wang, Yu, Jiang, and Chang}]{wang2018co}
Wang, S.; Yu, M.; Jiang, J.; and Chang, S. 2018.
\newblock A Co-Matching Model for Multi-choice Reading Comprehension.
\newblock In \emph{Proceedings of the 56th Annual Meeting of the Association
  for Computational Linguistics (ACL'18)}, 746--751.

\bibitem[{Wei et~al.(2022)Wei, Wang, Schuurmans, Bosma, Ichter, Xia, Chi, Le,
  and Zhou}]{DBLP:conf/nips/Wei0SBIXCLZ22}
Wei, J.; Wang, X.; Schuurmans, D.; Bosma, M.; Ichter, B.; Xia, F.; Chi, E.~H.;
  Le, Q.~V.; and Zhou, D. 2022.
\newblock Chain-of-Thought Prompting Elicits Reasoning in Large Language
  Models.
\newblock In \emph{Proceedings of the 36th International Conference on Neural
  Information Processing Systems (NeurIPS'22)}.

\bibitem[{Wu et~al.(2023)Wu, Fei, Qu, Ji, and Chua}]{wu2023nextgpt}
Wu, S.; Fei, H.; Qu, L.; Ji, W.; and Chua, T.-S. 2023.
\newblock NExT-GPT: Any-to-Any Multimodal LLM.
\newblock \emph{CoRR}, abs/2309.05519.

\bibitem[{Yao et~al.(2023)Yao, Yu, Zhao, Shafran, Griffiths, Cao, and
  Narasimhan}]{DBLP:journals/corr/abs-2305-10601}
Yao, S.; Yu, D.; Zhao, J.; Shafran, I.; Griffiths, T.~L.; Cao, Y.; and
  Narasimhan, K. 2023.
\newblock Tree of Thoughts: Deliberate Problem Solving with Large Language
  Models.
\newblock \emph{CoRR}, abs/2305.10601.

\bibitem[{Zeng et~al.(2023)Zeng, Wei, Liu, and Fu}]{DBLP:conf/acl/ZengWLF23}
Zeng, H.; Wei, B.; Liu, J.; and Fu, W. 2023.
\newblock Synthesize, Prompt and Transfer: Zero-shot Conversational Question
  Generation with Pre-trained Language Model.
\newblock In Rogers, A.; Boyd{-}Graber, J.~L.; and Okazaki, N., eds.,
  \emph{Proceedings of the 61st Annual Meeting of the Association for
  Computational Linguistics (ACL'23)}, 8989--9010. Association for
  Computational Linguistics.

\bibitem[{Zhang et~al.(2020)Zhang, Zhao, Wu, Zhang, Zhou, and
  Zhou}]{zhang2020dcmn+}
Zhang, S.; Zhao, H.; Wu, Y.; Zhang, Z.; Zhou, X.; and Zhou, X. 2020.
\newblock DCMN+: Dual co-matching network for multi-choice reading
  comprehension.
\newblock In \emph{Proceedings of the 34th AAAI Conference on Artificial
  Intelligence (AAAI'20)}, 9563--9570.

\bibitem[{Zhang et~al.(2023)Zhang, Zhang, Li, and
  Smola}]{DBLP:conf/iclr/0001Z0S23}
Zhang, Z.; Zhang, A.; Li, M.; and Smola, A. 2023.
\newblock Automatic Chain of Thought Prompting in Large Language Models.
\newblock In \emph{The Eleventh International Conference on Learning
  Representations, {ICLR} 2023, Kigali, Rwanda, May 1-5, 2023}. OpenReview.net.

\bibitem[{Zhao et~al.(2023)Zhao, Hu, Zhao, Shao, and
  Wang}]{DBLP:conf/acl/ZhaoHZSW23}
Zhao, Z.; Hu, L.; Zhao, H.; Shao, Y.; and Wang, Y. 2023.
\newblock Knowledgeable Parameter Efficient Tuning Network for Commonsense
  Question Answering.
\newblock In Rogers, A.; Boyd{-}Graber, J.~L.; and Okazaki, N., eds.,
  \emph{Proceedings of the 61st Annual Meeting of the Association for
  Computational Linguistics (ACL'23)}, 9051--9063. Association for
  Computational Linguistics.

\bibitem[{Zheng et~al.(2023{\natexlab{a}})Zheng, Ji, Li, Fei, Wu, Li, Li, and
  Teng}]{DBLP:journals/corr/abs-2306-03969}
Zheng, L.; Ji, D.; Li, F.; Fei, H.; Wu, S.; Li, J.; Li, B.; and Teng, C.
  2023{\natexlab{a}}.
\newblock {ECQED:} Emotion-Cause Quadruple Extraction in Dialogs.
\newblock \emph{CoRR}, abs/2306.03969.

\bibitem[{Zheng et~al.(2023{\natexlab{b}})Zheng, Li, Chai, Teng, and
  Ji}]{DBLP:conf/nlpcc/ZhengLCTJ23}
Zheng, L.; Li, F.; Chai, Y.; Teng, C.; and Ji, D. 2023{\natexlab{b}}.
\newblock A Bi-directional Multi-hop Inference Model for Joint Dialog Sentiment
  Classification and Act Recognition.
\newblock In \emph{Natural Language Processing and Chinese Computing - 12th
  National {CCF} Conference, {NLPCC} 2023, Foshan, China, October 12-15, 2023,
  Proceedings, Part {I}}, volume 14302, 235--248. Springer.

\end{thebibliography}
